\documentclass[10pt,twocolumn,letterpaper]{article}

\usepackage{wacv}              % To produce the CAMERA-READY version
\usepackage{graphicx}
\usepackage{amsmath}
\usepackage{amssymb}
\usepackage{booktabs}

\usepackage[pagebackref,breaklinks,colorlinks]{hyperref}

\usepackage[capitalize]{cleveref}
\crefname{section}{Sec.}{Secs.}
\Crefname{section}{Section}{Sections}
\Crefname{table}{Table}{Tables}
\crefname{table}{Tab.}{Tabs.}

\usepackage{multirow}
\newcommand{\figref}[1]{Fig.~\ref{#1}}
\newcommand{\tabref}[1]{Tab.~\ref{#1}}
\newcommand{\secref}[1]{Sec.~\ref{#1}}

\newcommand{\equref}[1]{Equ.~(\ref{#1})}

\def\eg{\emph{e.g.~}}

\def\etal{{\em et al.~}}
\def\sArt{{state-of-the-art~}}

\begin{document}

\title{THInImg: Cross-modal Steganography for Presenting Talking Heads in Images}

\author{Lin Zhao$^{1}\thanks{Corresponding author}$ \quad Hongxuan Li$^{1}$ \quad  Xuefei Ning$^2$ \quad Xinru Jiang$^3$ \\
$^1$TKLNDST, CS, Nankai University\\
$^2$Department of Electronic Engineering, Tsinghua University\\
$^3$Department of Computer Science, University of British Columbia\\
{\tt\small \{lin-zhao,hxli\}@mail.nankai.edu.cn; foxdoraame@gmail.com; xrjiang@student.ubc.ca}
}

% \and
\maketitle

%%%%%%%%% ABSTRACT
%%%%%%%%% ABSTRACT
\begin{abstract}
   Cross-modal Steganography is the practice of concealing secret signals in publicly available cover signals (distinct from the modality of the secret signals) unobtrusively. 
   While previous approaches primarily concentrated on concealing a relatively small amount of information, we propose THInImg, which manages to hide lengthy audio data (and subsequently decode talking head video) inside an identity image by leveraging the properties of human face, which can be effectively utilized for covert communication, transmission and copyright protection. 
   THInImg consists of two parts: the encoder and decoder. 
   Inside the encoder-decoder pipeline, we introduce a novel architecture that substantially increase the capacity of hiding audio in images. 
   %Taking the advantage of the generation capability, the recovered audio signals to synthesize talking head videos from cover images.
   %
   %With the above design, talking heads can be effectively hidden in identity images.
   %
   Moreover, our framework can be extended to iteratively hide multiple audio clips into an identity image, offering multiple levels of control over permissions.
   We conduct extensive experiments to prove the effectiveness of our method, demonstrating that THInImg can present \textbf{up to 80 seconds of high quality talking-head video (including audio) in an identity image with 160}\textbf{$\times$}\textbf{160 resolution}.
   %
    %Specifically, the input audio data is first converted into the mel-spectrogram and efficiently hidden into the THInImg in the encoder. 
   %
   %\red{The decoder extracts the speech information to the raw audio form and finally synthesizes the talking-head video (including audio) with the facial information from the THInImg. }
   %Therefore, different videos can be decoded according to the recovered audio clips with the corresponding levels of access control.
   %
\end{abstract}

%%%%%%%%% BODY TEXT
\section{Introduction}
Steganography is the art of discreetly embedding secret signals into overt signals, known as cover media, to ensure that only authorized recipients possess the capability to extract and decipher these concealed signals from the cover media ~\cite{hidden,steganogan, INN-hide, image-image, video-video}.
In the process of steganography, cross-modal steganography means that the secret signals and the cover signals have different modalities.
It is necessary for cross-modal steganography to unify the data from different modalities into a consistent format before concealing secret data, followed by the translation of the data format back to its original mode during the recovery process.
However, how to align the data of different modalities and embed a large amount of secret signals in cover signals inconspicuously remains a significant challenge nowadays.
%
%
%A facial image can be employed as a carrier to hide secret information, which makes it possible for us to hide text, audio, images and even other media information in it.
%
%
%while only using facial images as carriers cannot make the hidden information be dynamically expressed.

\begin{figure}[t]
\centering
\vspace{-1em}
\includegraphics[width=0.9\linewidth]
{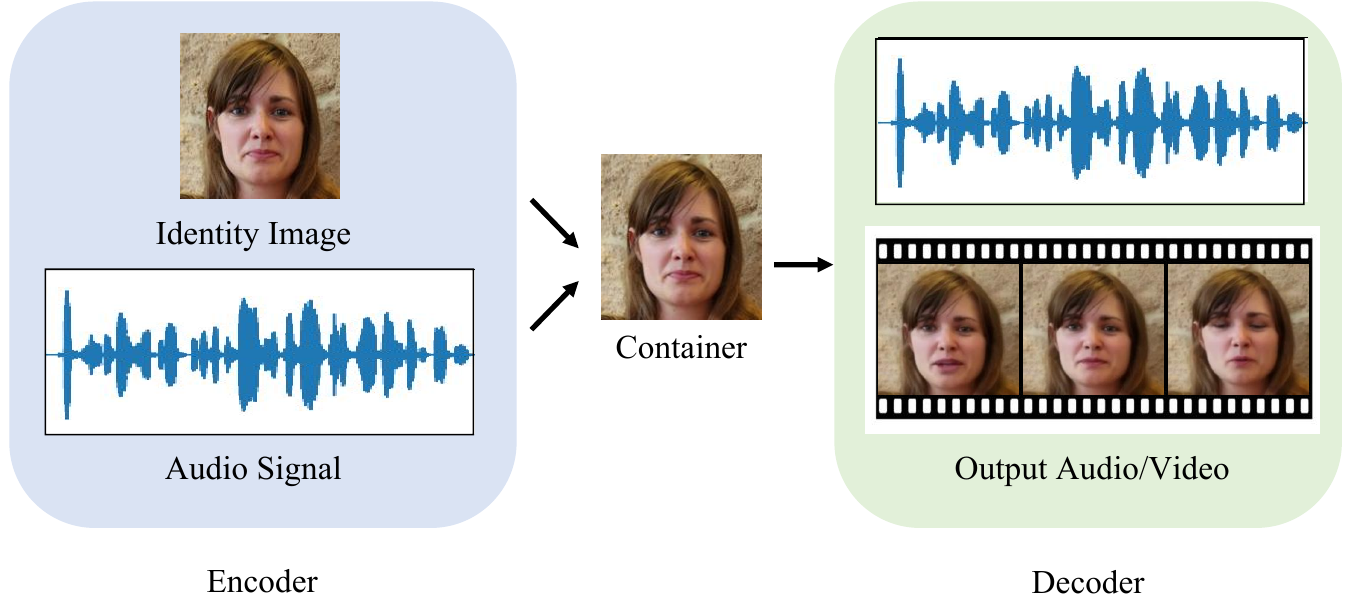}
\caption{The pipeline of THInImg to hide lengthy audio data in identity images and generate talking head videos (with audio) from the images.}
\label{OverAll}
\end{figure}

%There has been some exploration of the use of data from different modalities in cross-modal steganography, such as image, audio, and video~\cite{kishor2016review}. 
%
Cross-modal steganography encompasses various categories, and hiding audio in images is one of them. Several research efforts have ventured into exploring this aspect.
Traditional methods typically hide data in the spatial or transformation domains through manual design.
The least significant bits (LSB) algorithm, being the most commonly utilized among these methods, leverages the least significant bits of the cover media to conceal information.
Aagarsana \etal\cite{lsb} apply this algorithm to hide audio signals in RGB images in spatial domain.
Similarly, Hemalatha \etal\cite{wavelet} use this algorithm to hide audio signals in YCbCr images in the transform domain using integer wavelet transform.
However, as the limited number of insignificant bits in the cover media, only a small number of bits can be utilized when using the LSB algorithm.
%However， given that the LSB algorithm is limited to utilizing only the least significant bits in the cover image for hiding information, the available number of bits is significantly smaller in comparison to the entire image.
%
Consequently, the effective hiding capacity of secret information remains considerably limited.
Recently, some deep learning methods have been proposed in cross-modal steganography to improve the hiding effect.
For instance, Huu \etal\cite{auidoinimage1} achieve embedding a 4 second audio clip with short-time fourier transform (STFT) format in a 255$\times$255 image by using deep convolutional neural network (DCNN) model.
Gandikota \etal\cite{audioinimage2} utilize the generative adversarial networks (GAN) model to hide a 2 second audio clip in a 128$\times$128 image.
% Geleta \etal\cite{geleta2022pixinwav} use short-time discrete cosine transform to alter the format of audio to achieve embedding a 256$\times$256 image in a 1.53 second audio clip.
%
%Yang \etal\cite{yang2019hiding} conceal a 1 second 128$\times$128 color video in 1 second audio by encoding and decoding the video in the process.
%
It is clear that the length of concealed audio in the aforementioned works are not enough. This is because of two primary reasons. First, audio data typically has a larger size compared to image data. For instance, the bit count of a 20-second raw audio clip is approximately 2.33 times that of a $160 \times 160$ image.
Second, the human auditory system (HAS) demonstrates a heightened sensitivity to fluctuations in audio frequencies and the presence of noise. Therefore, the concealment and recovery of lengthy audio data are difficult.
%
%Considering the vast volume of audio-visual data, there remains significant value in exploring methods to conceal it in restricted data resource.

%Furthermore, naturally looking talking faces can be synthesized~\cite{talking-face1, talking-face2, talking-face3, talking-head5} when providing dynamic facial images and audio clips.
%
%

%3) We wish to propagate the audio-visual information secretly, and only authorized people can decode the corresponding dynamic audio-visual information.
%

In this paper, with the aim of expanding the generalization and usefulness of our system, we not only focus on hiding and recovering lengthy audio data in images, but also extend to decode human talking heads. 
In this way, our system can be applied in various scenarios, including but not limited to:
1) We enable covert video communication among multiple people/parties, ensuring privacy and content confidentiality. Furthermore, with solely image transmission instead of video, our system reduces video communication's bandwidth demands.
%
% 2) For secret communication, different individuals perceive the same facial image, but they can observe distinct videos.
2) Visually indistinguishable facial photos can be personalized for different viewers by using different decoders, like \eg "working" or "leisure" themed videos.
3) We can embed copyright information into images to provide evidence of image source and ownership.
%To be more specific, people can use image to transmit audio-visual information.
%
As shown in \figref{OverAll}, our system consists of encoder and decoder for encoding and decoding the audio-visual information.
%
%In the encoder, we hide lengthy audio data just in an image.
%
%It is noteworthy that we propose two innovative ways to minimize the data requirements for embedding.
By using vast prior knowledge of human face, talking head generation methods~\cite{talking-head3d, talking-face3, talkinghead} can obtain vivid talking head videos by an identity image and speech.
%
%by taking the advantage of existing talking head generation methods~\cite{talking-head3d, talking-face3, talkinghead}, which can obtain vivid talking head videos of the person by an identity image and speech. 
We apply a talking head generation model in the decoder to take the advantage of this characteristic.
To increase the hidden audio capacity, we innovatively propose a hiding-recovering architecture, which compressing data during the steganography process, and the compressed data (acoustic features) approach to \textbf{2/7} of its original size. 
This compression employs non-uniform techniques, aligning with human non-linear auditory perception. Consequently, while increasing capacity, audio quality restoration is ensured.

% Inside the encoder-decoder pipeline, we propose a hiding-recovering architecture that uses an innovative approach to compress data size during steganography to \textbf{2/7} of its original, thereby achieving a substantial increase in the capacity for concealing audio data.
% %
% Since using mel-spectrogram can enhance the perceptual fidelity of the recovered audio, the acoustic mel-spectrogram features compressed from the raw audio are used as secret information in the whole audio steganography process.
% %
%
%Therefore, as shown in \figref{OverAll}, during the encoding process, we only need to conceal audio signals in an identity image.
%In this way, the amount of data that needs to be hidden is greatly reduced.
%
 %the talking-head videos (plus audio) can be presented by just a single identity image.

To support multiple access levels in our system, we further propose to hide various audio clips in the image iteratively, which enables the reconstruction of diverse talking-head videos and their corresponding audio at varying access levels.
Extensive experiments are conducted on both single-speaker and multi-speaker databases to demonstrate the effectiveness of our system.
The results indicate that each container image can be decoded to different lengths (up to 80 seconds) of high quality videos with audio.

In summary, our main contributions are as follows:
\begin{itemize}
\item To the best of our knowledge, we are the first to hide lengthy audio data (and subsequently decode talking head video) inside an image.
\item We present a hiding-recovering architecture to significantly increase the capacity of hiding audio in images, enabling our THInImg to decode high-quality videos (including audio) of various lengths, up to 80 seconds for each image.
%\item We propose a new framework to hide speech in images, which can recover about 80 seconds of high-quality speech for each facial image. 
\item Our THInImg system provides support for multiple access levels, enabling the iterative embedding of multiple audio clips into a single image, different talking-head videos can be decoded at each level.
\end{itemize}

\section{Related Work}
\begin{figure*}[t]
\centering
\vspace{-1.3em}
\includegraphics[width=0.85\linewidth]{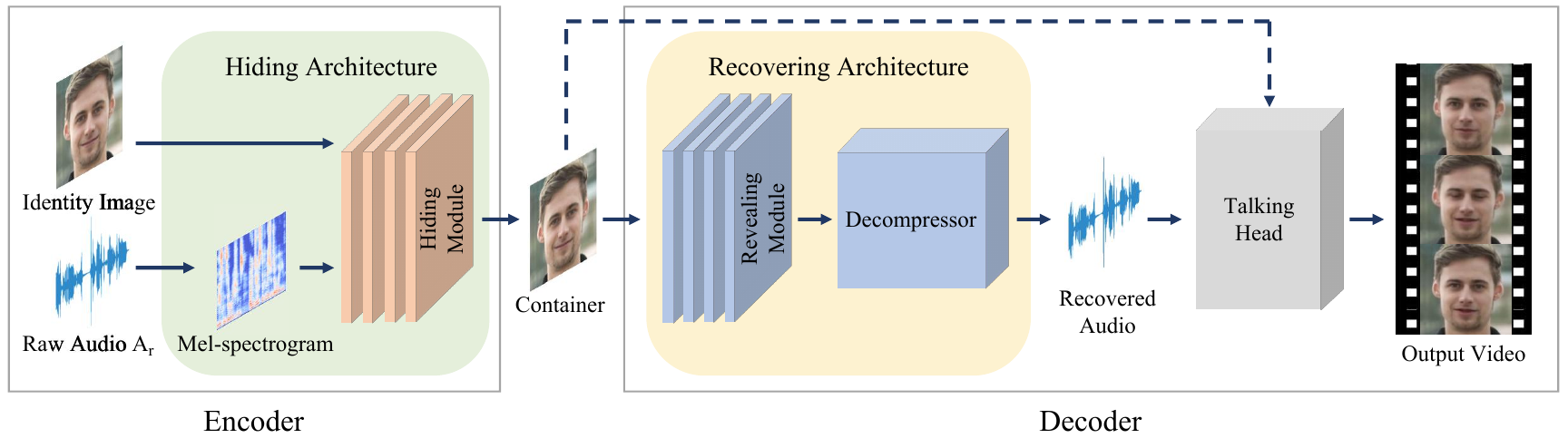}
\caption{The overall framework of THInImg. The encoder generates container images. Talking head videos are generated in the decoder.}
\label{ALL_arch}
\end{figure*}

\subsection{Cross-Modal Steganography}
In recent years, 
%the field of steganography has gained significant attention within the realm of information security.
%
numerous steganography methods have emerged, enabling the concealment of secret information in a diverse range of data types such as images, videos, and audio.
%
%By taking advantage of the diverse types of media, it becomes possible to embed the information deliberately.
%
Among them, images are the most commonly used ones that can be used as cover media~\cite{image-image,S_image,S_image2}.
Researchers have been able to embed many different forms of information in cover images.
For example, Tancik \etal\cite{stegastamp-image} introduce a learned steganographic algorithm to enable hiding hyperlink bit-strings in images.
Watermarks are discreetly embedded in images to safeguard copyright, and they find extensive usage in interactive mobile applications~\cite{water-image1,water-image2}.
%
%Besides that, watermarks can also be seen as a form of image steganography.
%
%Some watermarking methods are used in some interactive mobile phone applications~\cite{water-image1,water-image2} because of their significant practical potential.
%
Moreover, the video data is frequently favored as a kind of cover media for cross-modal steganography due to its substantial size and the statistical intricacy of its diverse features~\cite{data-video}.
For example, Wengrowski \etal\cite{light-video} develop a deep photographic steganography network to obscure light field messaging in the video.
Besides, Lu \etal\cite{lu2010effective} hide an image set in videos to protect the biometric data during transmission for secure personal identification.
Noteworthy, various audio communication solutions have been widely applied in the industry, making audio data a suitable type of cover media.
Cui \etal\cite{image-audio} propose a framework to obscure images into audio, which is the first to link both image and audio media in steganography.
Yang \etal\cite{yang2019hiding} first accomplish the concealment of videos in cross-modal steganography by compressing the video data and embedding it into audio signals.
In this paper, we hide talking heads in images by using human-face properties, which is first to embed lengthy  audio-visual information in cover .

\subsection{Audio-Driven Talking Head Generation}
With the rapid development of deep learning techniques, many approaches to generating audio-driven talking head have been introduced in recent years~\cite{talking-face3,talking-headv1,talking-headv2,talking-headv3,talking-headv4,talking-face2,talking-face1,talking-head5}.
%
%According to their purposes, we can briefly divide these methods into two categories.
%

At the early stage of the talking head study, researchers focus on generating talking-face videos by driving face regions cropped in video frames.
For instance, Chung \etal\cite{talking-face3} first suggest to create a video of the target face lip-synced with the audio.
However, this idea does not consider the time-dependency across video frames, resulting in abrupt lip movements.
After that, Song \etal\cite{talking-face2} incorporate the time-dependency of image and audio features in the recursive units of their generation network.
%
%Moreover, to improve the realistic effects, a novel cascade network structure is applied by Chen \etal\cite{talking-face1}, which leverages landmarks as intermediate results.
%
Nevertheless, the methods illustrated above consider only the face area, without the head movement and background, which makes the generated video unrealistic.

Later, to improve the realism of the generated video, the video generation process encompass not only the facial region but also the person's neck, hair, and the background.
For example, Suwajanakorn \etal\cite{talking-head5} focus on lip synthesis first then embed it into the original video to get a real video of Obama.
Indeed, the model's usability is constrained to a particular individual and it necessitates a substantial amount of training data in the form of long videos specific to that person.
To address this limitation, Zhou \etal\cite{talkinghead} and Yi \etal\cite{talking-headv2} develop more generalized models by integrating landmarks and 3D facial data, respectively. In our work, we apply the audio-driven talking head generation model to decode videos from the identity image.
%
%In addition, Ji \etal\cite{talking-headv1} consider the emotional features present in the audio when synthesizing talking-head videos to make the video more realistic and vivid.

Recently, there are also some works about talking-head video compression~\cite{talkinghead-add1, talkinghead-add2, talkinghead-add3, talkinghead-add4}. Cleverly leveraging generative models, Wang \etal\cite{talkinghead-add2} only needs to store a few frames and the key facial landmark information for each frame during the transmission process. 
Furthermore, video interpolation and super-resolution techniques are incorporated at the receiving end to further reduce the required number of bits \cite{talkinghead-add3}.
In contrast to them, we not only reduce the video bit rate to the size of an image, but also enable covert communication.

\section{Proposed Method}
\subsection{Encoder-Decoder}
Our THInImg can be seen as a new cross-modal steganography method to encode lengthy audio and decode talking head videos solely in an identity image.
As shown in \figref{ALL_arch}, THInImg consists of encoder and decoder.

\noindent\textbf{Encoder.} %
In the encoder, our primary purpose is to generate the container image ${I_c}$ by hiding the speech content in the identity image ${I_i}$.
%
%During the encoding process, we need to ensure that the visual effect of ${I_c}$ is consistent with ${I_i}$.
%
Formally, the encoder of our THInImg system can be expressed as:
\begin{equation}
\begin{split}
&{I_c} = E({A_r}, {I_i}), \\
\end{split}
\label{equ:T}
\end{equation}
where ${E( \cdot )}$ denotes the operation of the encoder, and ${A_r}$ is the raw audio.

\noindent\textbf{Decoder.} Decoding videos from a given container image requires two steps: recovering the audio signal and adapting a talking-head animating model ${G( \cdot )}$ to generate videos.
The formula is as follows:
\begin{equation}
\begin{split}
&{O_t} = {D}({I_c}) = {G}({Rec}({I_c})),
\end{split}
\label{equ:TR}
\end{equation}
where ${D( \cdot )}$ is the decoder, ${Rec( \cdot )}$ represents the recovering architecture that performs the recovering operation, and ${O_t}$ is the generated result video.
%In the encoder, we generate the container, while in the decoder, we extract the talking heads from the container.
%
In details, we apply the architecture in \cite{talkinghead}, which is the \sArt image-driven generation approach available, to produce plausible talking-head animations with facial expressions and head motions.
%

%
%Our pipeline consists of two parts: the encoder and the corresponding decoder.
%Besides, to train THInImg more effectively, we introduce an embeding-revealing module in \secref{sec:sec3}.
%
%Furthermore, we extend THInImg to support for multiple access levels, and the details are in \secref{sec:sec4}

%%%%%%%%%%%%%%%%%%%%%%%%%%%%%%%%%%%%%%%%%%%%%%%%% 
\subsection{The Hiding-Recovering Architecture}\label{sec:sec1}
As depicted in \figref{ALL_arch}, we propose a hiding-recovering architecture to efficiently hide lengthy audio data, including compression-decompression of the raw audio and embedding-revealing of the compressed acoustic features.
In the process, we need to compress the raw audio before embedding and decompress it after revealing.

\noindent\textbf{Audio compression-decompression.} Most existing audio-related steganography models either directly reshape the raw audio data into audio tensors~\cite{image-audio} or use the short-time Fourier transform (STFT) to calculate audio tensors~\cite{audio-audio,speech-ste}.
We innovatively apply non-uniform compression ${Com( \cdot )}$ to the audio, aligned with the non-linear auditory perception of humans, to significantly reduce the amount of data that needs to be hided while ensuring minimal audio quality loss.
By this way, we choose the Mel-spectrogram be the compressed acoustic features, which can better map the human auditory perception~\cite{mel}.
%We innovatively propose to convert raw audio ${V_r}$ into mel-spectrogram to enable the speech decoded from the THInImg to be without noise.
% %
% Most importantly, this processing can massively reduce the audio size, and thus the length of audio embedded in each container can be expanded by multiple times.
%
After getting the Mel-spectrogram, by splitting and stitching it into $c$ channels (the calculation is in \secref{sec:4.3}), we get ${A_s}$ as the input to be embedded, in which the length and width are consistent with the identity image. Thus the formula of audio compression is:
\begin{equation}
\begin{split}
&{A_s} = Comp({A_r}). \\
\end{split}
\label{equ:T}
\end{equation}

Obviously, when obtain the revealed audio signal ${A_{r_R}}$ from the container image ${I_c}$,
we need to decompress the raw audio waveforms ${A_{e}}$ from the Mel-spectrogram, and apply the DCNN model - WaveNet vocoder~\cite{wavenet_vocoder} as decompressor. Similarly, the formula of audio decompression is:
\begin{equation}
\begin{split}
&{A_e} = Decomp({A_{r_R}}). \\
\end{split}
\label{equ:T}
\end{equation}
\begin{figure}[t]
\centering
\vspace{-1.5em}
\includegraphics[width=1.0\linewidth]{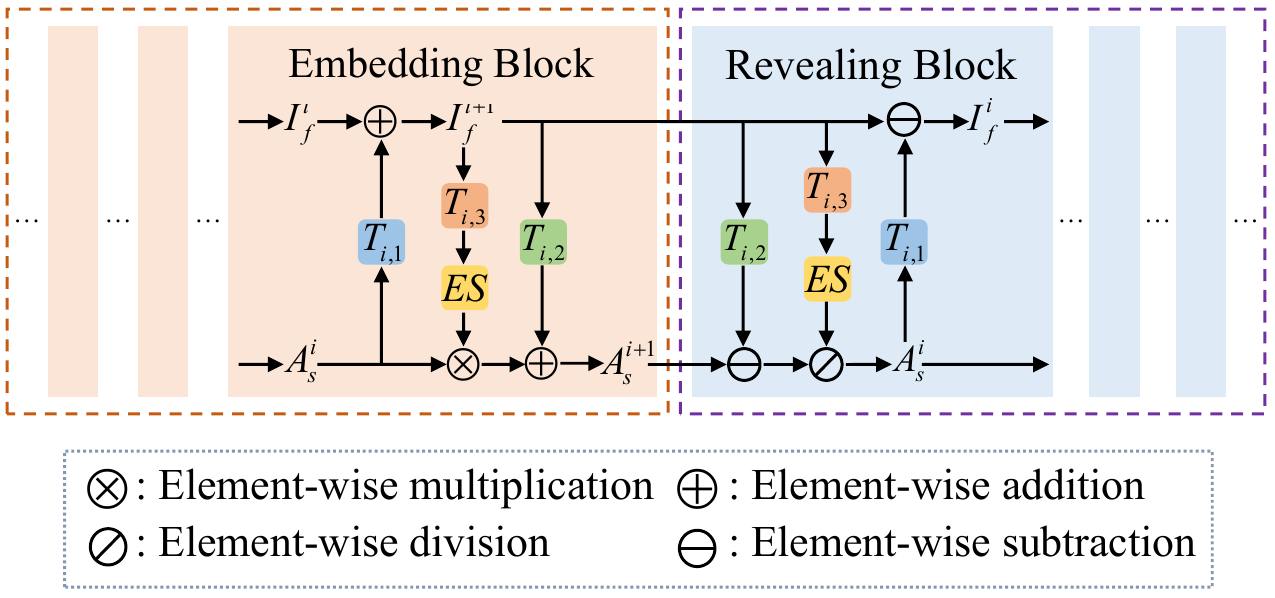}
\caption{The detail composition of one of 8 embedding-revealing blocks in our uniform module.}
% \caption{Our uniform module consists of 8 hiding-revealing blocks. We give in detail the exact composition of one of the blocks.}
\label{INN_block}
\end{figure}

\noindent\textbf{The Embedding-revealing Module. }\label{sec:sec2}
We introduce an invertible neural networks (INN) based network~\cite{INN-glow,INNguided,INN-SR,INN-hide,INN_video,INN_signal,INN_wave,INN_enhance} to embed and reveal $A_s$ uniformly, which has been demonstrated for efficient image processing~\cite{INN-hide,INN-glow,INN_enhance}.
Our embedding and revealing module is invertible, and it conducts both forward and backward propagation operations synchronously for enabling efficient training.
When the module is propagating forward, we embed ${A_s}$ in ${I_i}$ to output ${I_c}$ as \equref{equ:T}.
Similarly, when the module is propagating backward, we reveal the Mel-spectrogram ${A_{s_R}}$ and the facial image ${I_{i_R}}$ from ${I_c}$:
\begin{equation}
\begin{split}
&({{A_{s_R}}}, {{I_{i_R}}}) = {Rev}({I_c}),
\end{split}
\label{equ:TR}
\end{equation}
where ${Rev( \cdot )}$ represents the module that performs the revealing operation.
In more detail, the INN-based module consists of several invertible embedding-revealing blocks (8 in our method). The specific structure of each block is shown in \figref{INN_block}. When the $i$-th block in our network propagates forward, it performs the operations in the embedding block:
\begin{equation}
\begin{split}
&{{I_i}^{i+1}} = {I_i}^{i} + {{E_{i,1}}({A_s}^{i}}), \\
&{{A_s}^{i+1}} = {{A_s}^{i}}*ES({E_{i,3}}({I_i}^{i+1})) + {{E_{{i},2}}({I_i}^{i+1})},
\end{split}
\end{equation}
where ${A_s}^{i}$ and ${I_i}^{i}$ are the input of the $i$-th block. ${E_{i,j}}( \cdot )$ represents the $j$-th ($j=1, 2,\mbox{ or }3$) encoding module in the $i$-th block, which can be any form of neural network architecture. In our experiment, ${E_{i,k}}$ is the residual block.
%
%We apply the dense block~\cite{densenet} as the encoding model (note that such block can also be adapted by other specified networks).
%
${ES}( \cdot )$ refers to the sigmoid function followed by the exponent. 
We use ${ES}( \cdot )$ as a multiplier to strengthen the encoding ability. Likewise, when propagating backward, the network performs the operations in the revealing block:
\begin{equation}
\begin{split}
&{A_s}^{i} = {\frac{{{A_s}^{i+1}} - {{E_{i,2}}({I_i}^{i+1})}}{ES(E_{i,3}({I_i}^{i+1}))}}, \\
&{I_i}^{i} = {{I_i}^{i+1}} - {{E_{i,1}}({A_s}^{i})}.
\end{split}
\end{equation}
Here ${A_s}^{i+1}$ and ${I_i}^{i+1}$ are the input of the $i$-th block in the backward operations.

%\noindent\textbf{Hiding module.} In our work, we design a uniform module to hide and reveal ${A_s}$, making the solution much easier to train.
%
%More details on the composition of the module are presented in~\secref{sec:sec3}.

%

%
%
%Note that WaveNet shares several variants~\cite{wavenet1,wavenet2,wavenet3} when adopted by different applications.
%
%The WaveNet vocoder takes acoustic features into account when synthesizing speech, when using the speaker identity embedding as an input, the model can generate multi-person speech audio.
%Noteworthy, ${ID}$ is an significant tag when training the WaveNet vocoder for multi-person speech synthesis.
%
%With a proper identity, we can ensure that the facial and acoustic features decoded from our THInImg are consistent, in terms of the gender and the age.
%
%Therefore, it is well suited to our task.

\begin{figure}[t]
\centering
\vspace{-1.5em}
\includegraphics[width=0.8\linewidth]{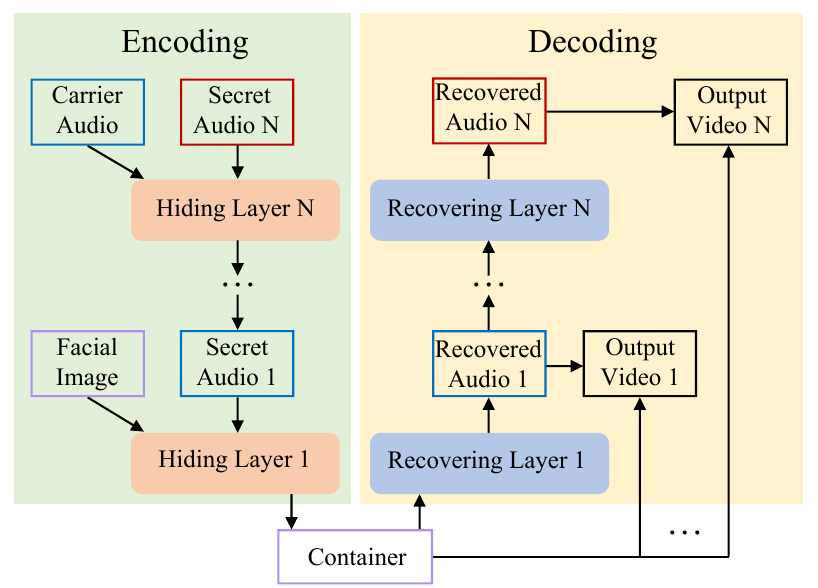}
% \caption{The diagram of nested embedding architecture. The figure is performing $N$ iterations, with the left part for encoding and the right for decoding, and the same color boxes mean same contents.}
\caption{The diagram of nested embedding architecture with $N$ iterations. The left part is for encoding and the right is for decoding, and the same color boxes mean same contents.}
\label{double_arch}
\end{figure}

\begin{figure*}[t]
\centering
\vspace{-1em}
\includegraphics[width=0.95\linewidth]{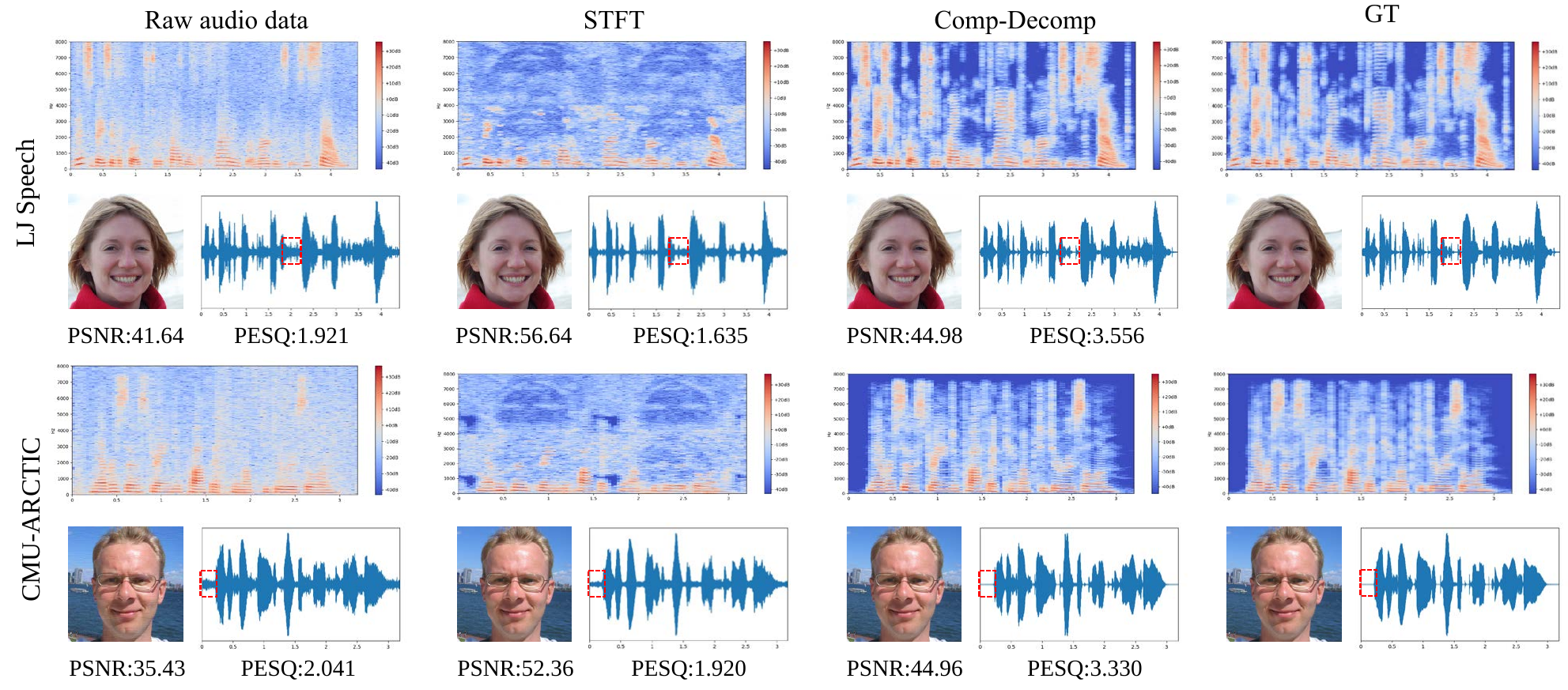}
\caption{The visual results of the container image and recovered audio from different methods. "GT" represents reference audio and the original identity image. From the red dashed box, only the results with audio compression-decompression avoid generating excess noise.}
\label{com_result}
\end{figure*}

%%%%%%%%%%%%%%%%%%%%%%%%%%%%%%%%%%%%%%%%%%%%%%%%%%%%%%%%%%%%%%%%%
\subsection{Nested Embedding Architecture}\label{sec:sec4}
As shown in \figref{double_arch}, we can cascade multiple embedding-revealing modules to form a nested embedding architecture, which enables the iterative hiding of various audio clips for giving different users different access levels.
To ensure the effectiveness of the entire system simultaneously, we adopt an end-to-end training approach for the architecture.
Here, we detail an example of a two-layer nested embedding architecture, which can be easily extended to multiple layers.
The first layer network ${E_1}$ serves the same purpose as the above base module in \secref{sec:sec2}, embedding the speech information of the first iteration ${A_{s_1}}$ into the identity image ${I_i}$ to get the container image ${I_c}$.
Moreover, ${A_{s_1}}$ acts as the container in the second layer network ${E_2}$ to embed the speech information of the second iteration ${A_{s_2}}$.
As before, the network is propagating forward during encoding:
\begin{equation}
\begin{split}
&{A_{s_1}} = {E_2}({A_{s_2}}, {A_c}), \\
&{I_c} = {E_1}({A_{s_1}}, {I_i}), \\
\end{split}
\label{equ:T_double}
\end{equation}
where ${A_c}$ represents the cover media of ${E_2}$. When ${I_c}$ needs to be decoded, the network conducts backward propagation:
\begin{equation}
\begin{split}
&({A_{{s_1}_R}}, {{I_{i_R}}}) = {E_{1_R}}({I_c}), \\
&({A_{{s_2}_R}}, {A_{c_R}}) = {E_{2_R}}({A_{{s_1}_R}}), \\
\end{split}
\label{equ:TR_double}
\end{equation}
where ${A_{{s_1}_R}}$ and ${A_{{s_2}_R}}$ denote the speech information decoded by the first and second iterations of ${I_c}$. ${E_{1_R}}$ and ${E_{2_R}}$ represent the recovering operation of the network during the decoding process, while ${I_{i_R}}$ and ${A_{c_R}}$ are the obtained cover media, respectively.

%%%%%%%%%%%%%%%%%%%%%%%%%%%%%%%%%%%%%%%%%%%%%
\subsection{Loss Function}
The purpose of the embedding-revealing module in \secref{sec:sec2} is twofold.
Firstly, both ${I_c}$ and ${I_{c_R}}$ need to be as similar as possible to the identity image ${I_c}$, so their loss functions can be expressed separately as:
\begin{equation}
\begin{split}
{\ell _C} = \frac{1}{N}\sum\limits_{i = 1}^N {{{\left\| {{I_c} - {I_i}} \right\|}^2}},
\end{split}
\end{equation}
\begin{equation}
\begin{split}
{\ell _I} = \frac{1}{N}\sum\limits_{i = 1}^N {{{\left\| {{I_{i_R}} - {I_i}} \right\|}^2}}.
\end{split}
\end{equation}
Here $N$ represents the number of the training images.
In addition, the ${A_{s_R}}$ recovered from ${I_c}$ should be consistent with the original ${A_s}$, which can be expressed by the following formula:
\begin{equation}
\begin{split}
{\ell _A} = \frac{1}{N}\sum\limits_{i = 1}^N {{{\left\| {{A_{s_R}} - {A_s}} \right\|}^2}}.
\end{split}
\end{equation}
Therefore, the final training loss is defined as a weighted sum of the losses mentioned above:
\begin{equation}
\begin{split}
{\ell _{tr}} = {\lambda _C}{\ell _C} + {\lambda _I}{\ell _I} + {\lambda _A}{\ell _A},
\end{split}
\end{equation}
where ${\lambda _C}$, ${\lambda _I}$ and ${\lambda _A}$ refer to the balanced weights. To obtain the container image ${I_c}$ and recovered Mel-spectrogram ${A_{s_R}}$, we set ${\lambda _C} = {\lambda _A} = 32$, ${\lambda _I} = 1$ during training.

The architecture in \secref{sec:sec4} contains two embedding-revealing modules, both of which employ the loss function described above and perform end-to-end training. Therefore, the loss of the architecture during training can be further expressed as:
\begin{equation}
\begin{split}
{\ell _{tr}} = {\ell _{tr_1}} +{\ell _{tr_2}},
\end{split}
\end{equation}
where ${\ell _{tr_1}}$ and ${\ell _{tr_2}}$ represent the loss of the two embedding-revealing modules, respectively.

\begin{table*}
\begin{center}
\small
\begin{tabular}{c|c c|c c|c c|c c}
\hline
Database & \multicolumn{4}{c|}{LJ Speech} & \multicolumn{4}{c}{CMU-ARCTIC} \\
\hline
\multirow{2}{*}{Form$\backslash$0$\sim$10s} & \multicolumn{2}{c|}{Container image} & \multicolumn{2}{c|}{Extracted video} & \multicolumn{2}{c|}{Container image} & \multicolumn{2}{c}{Extracted video} \\
\cline{2-9} & PSNR & SSIM & PESQ & PIQE & PSNR & SSIM & PESQ & PIQE \\  
\hline
Raw audio & 41.50 & 0.979 & 1.855 & 46.87 & 38.08 & 0.965 & 2.026 & 47.53 \\
\hline
STFT & \textbf{55.94} & \textbf{0.999} & 1.697 & 46.77 & \textbf{53.20} & \textbf{0.998} & 2.132 & \textbf{46.23}\\
\hline
Comp-Decomp & 41.50 & 0.979 & \textbf{3.639} & \textbf{46.59} & 44.66 & 0.989 & \textbf{3.215} & \textbf{46.23} \\
\hline
\end{tabular}
\end{center}
\caption{Quantitative comparison with different methods, audio lengths ranging from 0$\sim$10s for each image.}
\label{tab:com_result}
\end{table*}

\begin{figure*}[t]
\centering
%\vspace{1.0em}
\includegraphics[width=0.85\linewidth]{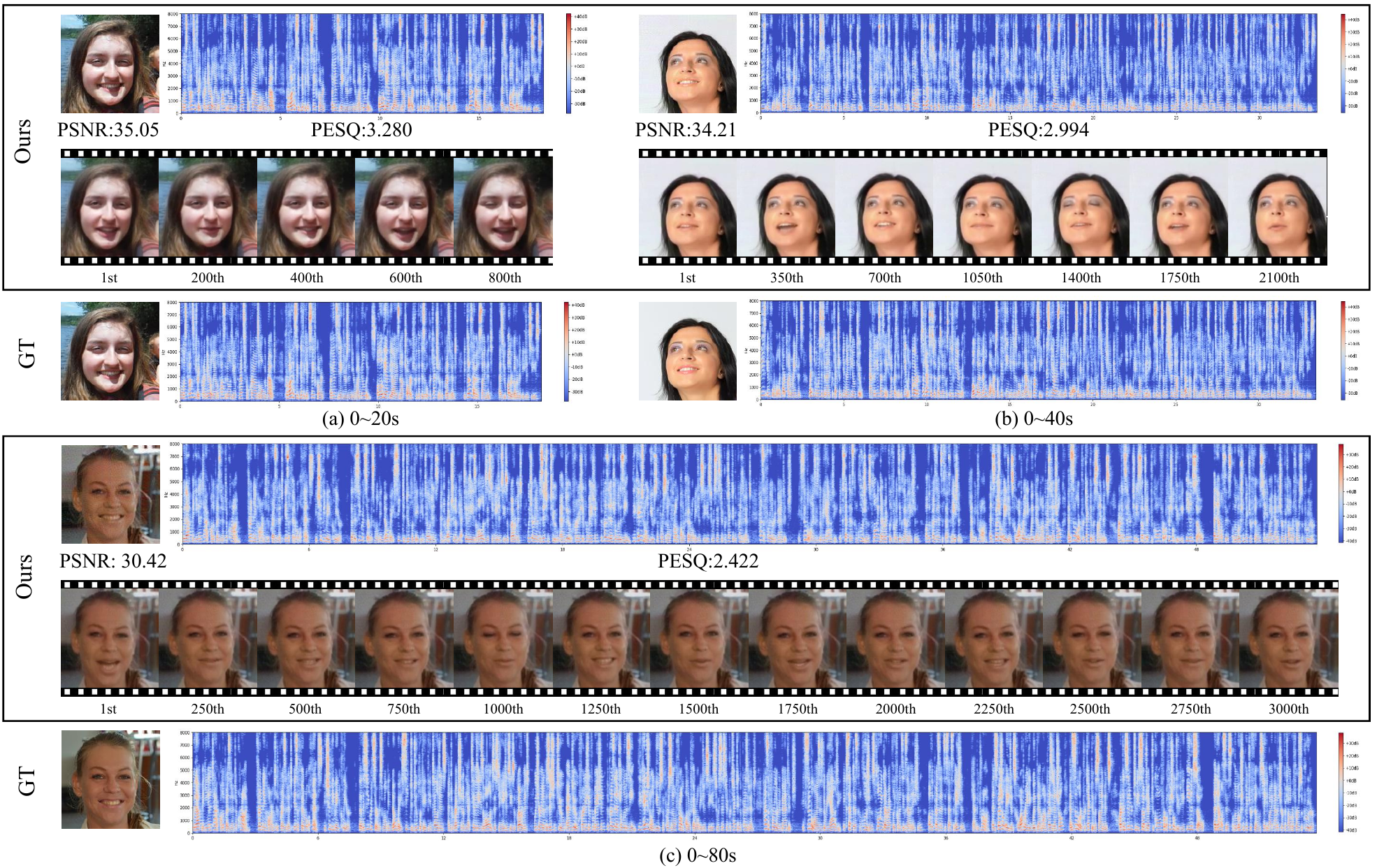}
\caption{The decoded audio and video frames corresponding to different audio lengths in the LJ Speech database.}
\label{ab_result_lj}
\end{figure*}

% \begin{figure*}[t]
% \centering
% \vspace{0.5em}
% \includegraphics[width=0.98\linewidth]{figures/ResultDouble-1112.pdf}
% \caption{The decoded audio and video frames corresponding to different audio lengths in the CMU-ARCTIC database. More results of both datasets are in the supplementary material.}
% \label{ab_result_cmu}
% \end{figure*}

\begin{figure}[t]
\centering
%\vspace{0.5em}
\includegraphics[width=0.7\linewidth]{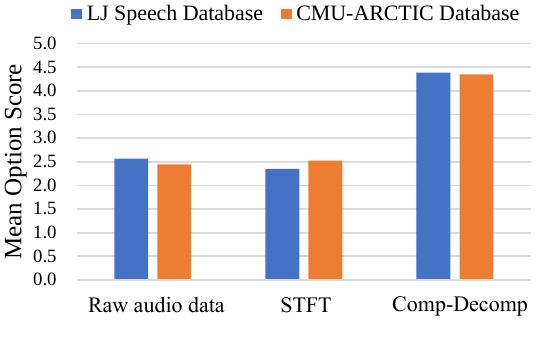}
% \caption{Comparison of MOS values among different methods. We conduct separate statistics for the scores of the generated talking-head videos of the two databases.}
\caption{MOS values comparison across methods, with separate statistics for generated talking-head videos from two databases.}
\setlength{\abovecaptionskip}{2cm}
\label{user}
\end{figure}

\begin{table*}
%\small
\begin{center}
\small
\vspace{-1em}
\begin{tabular}{c|c c|c c|c c|c c}
\hline
Database & \multicolumn{4}{c|}{LJ Speech} & \multicolumn{4}{c}{CMU-ARCTIC} \\
\hline
\multirow{2}{*}{Time range} & \multicolumn{2}{c|}{Container image} & \multicolumn{2}{c|}{Extracted video} & \multicolumn{2}{c|}{Container image} & \multicolumn{2}{c}{Extracted video} \\
\cline{2-9} & PSNR & SSIM & PESQ & PIQE & PSNR & SSIM & PESQ & PIQE \\  
\hline
0$\sim$20s & 39.72 & 0.970 & 3.370 & 46.68 & 40.44 & 0.976 & 3.107 & 46.43 \\
\hline
0$\sim$40s & 30.85 & 0.895 & 3.111 & 48.14 & 37.00 & 0.954 & 2.880 & 47.18  \\
\hline
0$\sim$80s & 28.25 & 0.811 & 2.482 & 45.92 & 30.03 & 0.893 & 2.382 & 49.23 \\
\hline
\end{tabular}
\end{center}
\caption{Quantitative results of embedding different ranges of audio lengths in the THInImg system.}
\label{tab:ab_result_one}
\end{table*}

\begin{figure*}[t]
\centering
% \vspace{0.5em}
\includegraphics[width=0.9\linewidth]{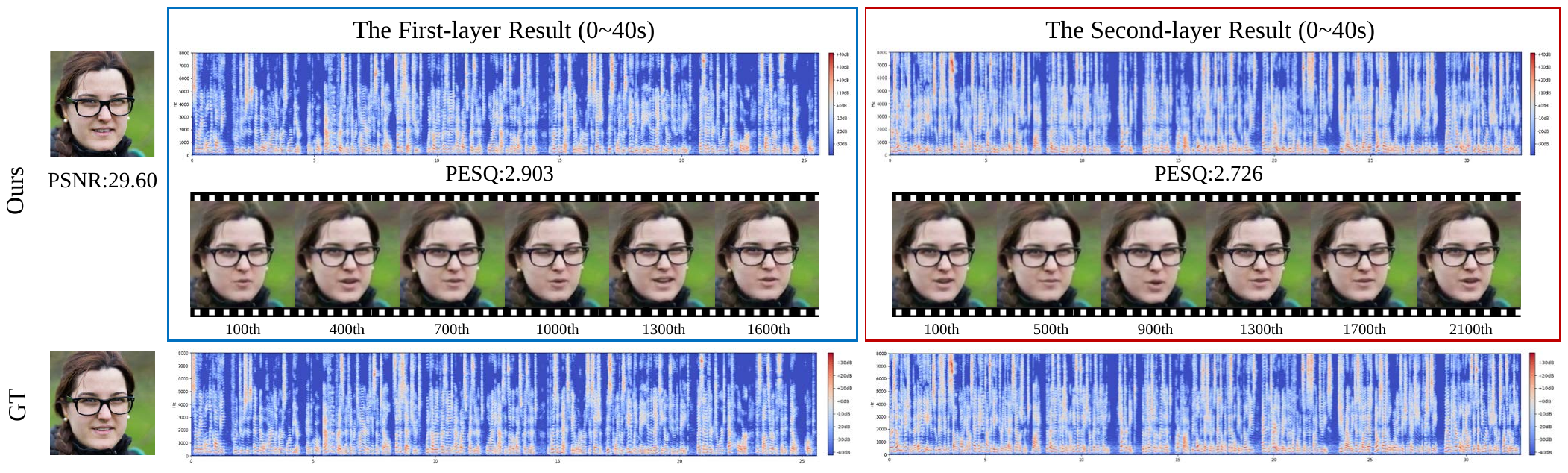}
\caption{Visual results of the double-layer nested embedding architecture in hiding 0$\sim$80s audio. Results for other audio length ranges are in the supplementary material.}
\label{ab_result_double}
\end{figure*}

\begin{table*}
\small
\begin{center}
\begin{tabular}{c|c c|c c|c c}
\hline
Database & \multicolumn{6}{c}{LJ Speech} \\
\hline
\multirow{2}{*}{Time range} & \multicolumn{2}{c|}{Container image} & \multicolumn{2}{c|}{First video} & \multicolumn{2}{c}{Second video} \\
\cline{2-7} & PSNR & SSIM & PESQ & PIQE & PESQ & PIQE \\  
\hline
0$\sim$20s & 39.60 & 0.967 & 3.521 & 46.44 & 3.414 & 46.58 \\
\hline
0$\sim$40s & 36.32 & 0.938 & 3.424 & 47.42 & 3.280 & 47.56 \\
\hline
0$\sim$80s & 30.87 & 0.864 & 2.912 & 49.58 & 2.857 & 49.63 \\
\hline
%0$\sim$80s & 27.74 & 0.812  \\
%\hline
\end{tabular}
\end{center}
\caption{Quantitative results of hiding different ranges of audio lengths in the THInImg. ``First video" and "Second video" mean the extracted talking head videos of the first layer and the second layer.}
\label{tab:ab_result_double}
\end{table*}

\section{Experiments}
\subsection{Datasets}
We use a facial image database, a single-person speech database, and a multi-person speech database in our experiments.

\noindent\textbf{FFHQ} The FFHQ Database~\cite{FFHQ} consists of $70{,}000$ high-quality identity images and contains considerable variation in terms of age, ethnicity, and image background.

\noindent\textbf{LJ Speech} The LJ Speech Database consists of $13{,}100$ short audio clips of passages from 7 non-fiction books read by a single speaker.
The audio clips vary in length from 1 to 10 seconds and have a total utterance duration of approximately 24 hours.

\noindent\textbf{CMU-ARCTIC} The CMU-ARCTIC Database~\cite{cmu_arctic} is a multi-person speech database. We use speech data of 7 speakers in the database: bdl, slt, jmk, awb, rms, clb, and ksp.
The total number of utterances is about $1{,}132$ per speaker, and the total length is about 1 hour per speaker.

\subsection{Metrics}
We employ Perceptual Evaluation of Speech Quality (PESQ)~\cite{PESQ} as a quantitative metric to objectively assess audio quality, which ranges from -0.5 to 4.5.
The Perception based Image Quality Evaluator (PIQE)~\cite{PIQE} is used to assess the generated talking-head video quality.
%
%We evaluate the generated talking-head video quality using video quality evaluation metric PIQE~\cite{PIQE}. 
%
The range of PIQE metric is 0 to 100, with lower values indicating higher video quality.
We incorporate Mean Opinion Score (MOS) to obtain human evaluations regarding audio-visual information quality. 
Peak Signal-to-Noise Ratio (PSNR) and Structural Similarity Index Metric (SSIM) metrics are utilized to evaluate the visual quality of container images.

%%%%%%%%%%%%%%%%%%%%%%%%%%%%%%%%%%%%%%%%%%%%
\subsection{Implementation Details}
We conducted two sets of experiments: One is using FFHQ images and audio clips from the single-person database LJ Speech, the training set consisted of 12,522 audio clips, and the testing set comprised 578 audio clips. Each paired with an equivalent number of images.
Similarly, the other is using equal number of images from FFHQ database and audio clips from multi-person speech database CMU-ARCTIC, a training set of 7,580 utterances and a testing set of 350 utterances are utilized. To ensure fairness, an equal number of utterances were randomly selected from each person in the CMU-ARCTIC database for training.
%
% We conducted two sets of experiments: one using FFHQ images paired with the single-person database LJ Speech audio, and the other using FFHQ images with the multi-person speech database CMU-ARCTIC.
% %
% For training single-person speech database, the training set consisted of 12,522 audio clips, and the testing set comprised 578 audio clips, all sourced from the LJ speech database.
% %
% For the multi-person speech experiment, a training set of 7,580 utterances and a testing set of 350 utterances are utilized. To ensure fairness, an equal number of utterances were randomly selected from each person in the CMU-ARCTIC multi-person speech database for training.
%
%

In addition, to verify effectiveness of the nested embedding architecture, we conduct experiments on LJ Speech database using a two-layer architecture as an example.
The first half of the training and testing sets are for the first layer network, and the second half is for the second layer.

\noindent\textbf{Input} \label{sec:4.2} 
%
% For each experiment, we apply the corresponding number of facial images in the FFHQ database and audio clips for training and testing.
%
%In order to adapt to the number of audio channels and the talking-head model
The images are resized to $160 \times 160$ resolution and the audio clips are adjusted to 16kHz.
For the processing of the audio, spectrum is represented by applying the STFT with $1{,}024$ FFT frequency bins and a sliding window with a shift 256. The number of Mel filters for compressing is 80.

\noindent\textbf{Training} 
The proposed algorithm is implemented in Pytorch~\cite{pytorch}, and an Nvidia 2080Ti GPU is used for acceleration. 
We train the decompressor for 2000 epochs in all experiments, while the embedding-revealing module and its double-layer nested hiding architecture for 100 epochs.
The ADAM optimizer is applied for all networks, while the learning rate is set to 2e-4 and 1e-2.
%

%%%%%%%%%%%%%%%%%%%%%%%%%%%%%%%%%%%%%%%%%%%%%%%%%%
\subsection{Performance Experiments}\label{sec:4.3} 
\textbf{Verification of audio compression-decompression effectiveness.} We demonstrate the effectiveness by comparing the hiding-recovering architecture with two methods without compression.
The architectures of two methods are applied with the raw audio data and STFT, respectively.
Between them, we regard the architecture utilizing raw audio data as the baseline. The architecture using STFT, which is applied in \cite{auidoinimage1, audioinimage2}, as a method to improve the baseline. 

To enable each image to obscure different lengths of speech, we randomly select the audio length corresponding to each image from 0 to 10s both during the training and testing processes.
We reshape each of the three formats of data into a tensor with the size of $h \times w \times c$, where $h = w = 160$, whose value is the same as the image size.
%
%To ensure that all tensors for each format of data are at the uniform length $h \times w \times c$, we pad them by concatenating zero tensors of the desired length with each input tensor of raw format and mel-spectrogram format.
%
%Suppose the length of the input tensor is $l$, then the length of the zero tensor concatenated to it is $(h \times w \times c-l)$.
%
The STFT spectrum is complex with a form of $a + bi$, so we need to embed both the real part $a$ and the imaginary part $b$.
Adjusting different window sizes of the STFT spectrum can control the size of the input tensor as $h \times w \times c$.
After the standardization, we get $c = 2$ for the Mel-spectrogram, $c = 7$ for the raw audio and $c = 4$ for the STFT spectrum.

\noindent\textbf{Analysis of quantitative results.}
The results on both single-speaker and multi-speaker databases are presented in \tabref{tab:com_result}.
We can see that only the results obtained using Mel-spectrogram as input achieve high scores for the recovered audio.
The container image and the generated talking-head videos obtained using these three audio formats can each achieve a satisfactory score.
%
%Since in the STFT format the speech does not need to be concatenated to zero tensor that causes relatively larger distortion to the image.
%
The PSNR and SSIM values of the STFT format are higher than the other two methods, but the values of its PESQ are unsatisfactory.

\noindent\textbf{Analysis of qualitative results.}
We present the visual results of container images and recovered audio in \figref{com_result}.
The visualization of the recovered audio includes its waveform plot in the time domain and its Mel-spectrogram plot in the frequency domain.
As can be seen, it is challenging to discover the difference between the container image of each method and the original facial image with the naked eye.
However, for the recovered audio, the waveform plots of the two methods -- using raw format or STFT format -- are different from that of the reference audio.
In addition, all waveforms are noisy except for the results obtained using the Mel-spectrogram as input, especially as shown by the dashed boxes \figref{com_result}, which represents gaps in speech.
Mel-spectrogram plots visually indicate that only the third column method produces similar output with reference audio.

\noindent\textbf{Naturalness MOS.}
To obtain the subjective mean opinion score (MOS) of each method, we randomly selected 2 audio clips per person in the CMU-ARCTIC database and 6 audio clips in the LJ Speech database to generate the videos, making a total of 20 talking-head videos for rating.
We invited 30 participants who rated the samples on a scale of 0$\sim$5 with 0.5 point increments.
The testers rate the talking-head videos by a combination of audio quality and video reality.
As shown in \figref{user}, the highest values are reached when using the Mel-spectrogram format in both databases.

%%%%%%%%%%%%%%%%%%%%%%%%%%%%%%%%%%%%%%%%%%%%%%%%%%%%%%
\subsection{Nested Embedding and Capacity Study}\label{sec:4.5} 
We test \textbf{the THInImg capacity for both the base and the double-layer nested embedding architectures}.
We splice the audio clips in two databases separately to perform experiments on embedding speech information of different ranges.
The quantitative results of the base single-layer model are shown in \tabref{tab:ab_result_one}.
As the audio lengths increase, the number of channels $c$ in the network increases, performance decreases for all metrics.
It can be seen more visually in \figref{ab_result_lj} for single speaker and \figref{ab_result_double} for multiple people that when audio lengths range is less than 40s, the results are satisfactory for both container images and recovered audio.
When the length reaches 80 seconds, although the audio is noiseless and the Mel-spectrogram plot is approximately the same, few artifacts in the container image.

The results are shown in \tabref{tab:ab_result_double} and \figref{ab_result_double} for the two-layer nested embedding architecture.
The audio ranges in the first and second layers are the same, and the audio quality in the deep layer is slightly worse than that in the shallow layer.
As mentioned above, the effect is satisfactory but worsens as the audio lengths increase.
We show that our THInImg can hide and reveal 80 seconds speech with guaranteed quality through experiments.
The recovered audio and the final generated videos of talking heads are displayed in the supplementary material.

\subsection{Discussion about number of nested layers}
We observe two interesting phenomena. 
First, through \tabref{tab:ab_result_one} and \tabref{tab:ab_result_double}, when the maximum audio lengths of hiding are 40s and 80s, the qualities of the image and audio in the two-layer nested architecture are better than the base single-layer model.
It is because when hiding same length audio, the two-layer nested architecture reduces concealed audio length in each layer, compared to the base single-layer model. Therefore, when concealing audio over a longer range, the two-layer nesting performs more effectively.
It can be inferred that models with higher nested layers tend to exhibit better performance when the nesting depths are in an appropriate range.
Second, as mentioned in the results of the nested architecture, the quality of deep layer is worse than shallow layer.
%Second, as the number of layers increasing, excessively high model loss results in notably poor quality, and the quality of deep layer is worse than shallow layer.
%
It is easily inferable that the quality will be degraded when the number of layers reaches a certain threshold due to the loss incurred from training a multi-layer architecture exceeds the benefits derived from less data embedded each layer mentioned in the first point.
As the number of layers continues to increase, excessively high model loss will result in notably poor quality. 

Considering the above two points collectively, the selection of the number of layers for network nesting is a topic worthy of discussion when hiding audio of a specific length.
In the future, we will explore several sub-questions arising from the aforementioned issue, including: 1. Selection of weights for different layers in multi-layer nesting. 2. The number of layers can be nested until quality severely degrades. 3. Regarding our system, what is the optimal number of layers for achieving the best results, and does this choice allow more data (longer than 80s) to be hidden?

\section{Conclusion}
This paper proposes THInImg, a cross-modal steganograply method to hide lengthy audio data and decode the talking-head videos with audio (up to 80 seconds) in 160x160 identity images.
There is an encoder and a decoder in THInImg for encoding audio and decoding the videos.
In the hiding-recovering pipeline, a novel architecture was introduced to simultaneously increase the length of concealed audio while ensuring audio quality.
%To achieve the concealment of a large amount of talking heads, we designed two ways to reduce the data footprint.
%
%First, with the help of talking head video generation models, we hide speech content instead of videos.
%
%Furthermore, we convert the raw audio into the mel-spectrogram, which effectively improves the quality of the recovered audio signal and significantly increases the length of the audio obscured in each image.
%
%In the proposed solution, the encoder embeds the audio into the facial image to get THInImg, and the decoder extracts it and adapts a talking-head model to generate the final video.
%
%In addition, a new framework has been proposed in the pipeline for embedding different lengths of speech in the image.
%
Furthermore, We extended the structure to allow for nested embedding, providing different access priorities for users.
Numerous experiments have been conducted to demonstrate the effectiveness of our method.
%
% The limitation of the work is that talking head videos are lossy compared to real video. In the future, we would like to consider how to further reduce this loss in our system.
%\newpage
%\endinput
%%
%% End of file `sample-authordraft.tex'.

{\small
\bibliographystyle{ieee_fullname}
\bibliography{egbib}
}

\end{document}